\pdfoutput=1
\documentclass[11pt]{article}
\usepackage{graphicx}
\usepackage{subfig}
\usepackage{ACL2023}

\usepackage{multirow}
\usepackage{times}
\usepackage{latexsym}
\usepackage{amsmath,amssymb,amsfonts} 
\usepackage[T1]{fontenc}

\usepackage[utf8]{inputenc}

\usepackage{microtype}

\usepackage{inconsolata}

\title{Enhancing Document-level Event Argument Extraction with \\ Contextual Clues and Role Relevance}
\author{Wanlong Liu\textsuperscript{\rm 1},
    Shaohuan Cheng\textsuperscript{\rm 1},
    Dingyi Zeng\textsuperscript{\rm 1},
    Hong Qu\textsuperscript{\rm 1}\thanks{~~Corresponding author: Hong Qu} \\
        {\textsuperscript{\rm 1}{University of Electronic Science and Technology of China, Chengdu, China}} \\ \{liuwanlong, shaohuancheng, zengdingyi\}@std.uestc.edu.cn,
        hongqu@uestc.edu.cn}

\begin{document}
\maketitle
\begin{abstract}
Document-level event argument extraction poses new challenges of long input and cross-sentence inference compared to its sentence-level counterpart.
However, most prior works focus on capturing the relations between candidate arguments and the event trigger in each event, ignoring two crucial points: a) non-argument contextual clue information; b) the relevance among argument roles.
In this paper, we propose a SCPRG (\textbf{S}pan-trigger-based
\textbf{C}ontextual \textbf{P}ooling and latent \textbf{R}ole
\textbf{G}uidance) model, which contains two novel and effective modules for the above problem. The \textbf{S}pan-\textbf{T}rigger-based
\textbf{C}ontextual \textbf{P}ooling (STCP) adaptively selects and aggregates the information of  non-argument clue words based on the context attention weights of specific argument-trigger pairs from pre-trained model.
The \textbf{R}ole-based
\textbf{L}atent \textbf{I}nformation \textbf{G}uidance (RLIG) module constructs latent role representations, makes them interact through role-interactive encoding to capture semantic relevance, and merges them into candidate arguments. Both STCP and RLIG introduce no more than 1\% new parameters compared with the base model and can be easily applied
to other event extraction models, which are compact and transplantable.
Experiments on two public datasets show that our SCPRG outperforms previous state-of-the-art methods, with 1.13
F1 and 2.64 F1 improvements on RAMS and WikiEvents respectively. Further analyses illustrate the interpretability of our model. 
\end{abstract}

\section{Introduction}

Event argument extraction (EAE) aims to identify the arguments of events formed as entities in text and predict their roles in the related event.  
As the key step of event extraction (EE), EAE is an important NLP task with widespread applications, such as recommendation systems~\cite{lietal2020gaia} and dialogue systems~\cite{Zhang2020.08.03.20167569} for presenting unstructured text containing event information in structured form.
Compared with previous works~\cite{liuetal2018jointly, waddenetal2019entity,  tongetal2020improving} focusing on sentence-level EAE, more and more recent works tend to explore document-level EAE~\cite{wangetal2022query, yangetal2021document, xu2022two}, which needs to solve long-distance dependency~\cite{ebner2020multi} and cross-sentence inference~\cite{lietal2021document} problems.
\begin{figure}[tbp]
    \centering
    \includegraphics[width=1.0\linewidth]{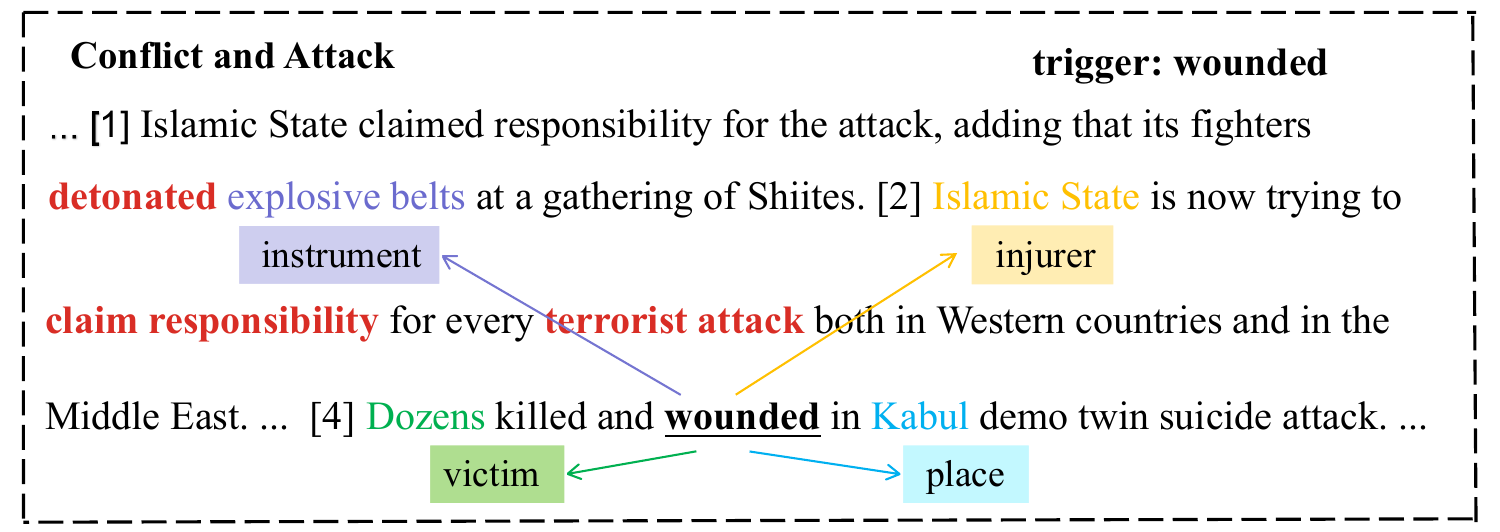}
    
    \caption{A document from RAMS~\cite{ebner2020multi} dataset. Event \textit{Conflict and Attack} is triggered by
\textit{wounded}, with four arguments of different roles scattering across
the document. Words in red are non-argument clue words meaningful for argument extraction.}
    \label{fig:fig1}
\end{figure}
 Therefore, many works~\cite{zhang2020two,  pouranbenveysehetal-2022-document} try to construct graphs based on heuristic rules~\cite{xuetal2021document} or syntactic structures~\cite{xu2022two} and model logical reasoning with Graph Neural Networks~\cite{kipf2016semi, 10096316}. 
However, all of state-of-the-art works ignore two crucial points: (a) the non-argument clue information; (b) the relevance among argument roles.

Non-argument clues are contextual text except target arguments that can provide important guiding information for the prediction of many complex argument roles. For example, in Figure~\ref{fig:fig1}, for the event \textit{Conflict and Attack}, non-argument clues \textit{detonated}, \textit{claim responsibility} and \textit{terrorist attack} can provide significant clue information for identifying arguments \textit{explosive belts} and \textit{Islamic State}. 
However, many previous works~\cite{lietal2021document, xu2022two} only utilize pre-trained transformer-based encoder to obtain global context information implicitly, ignoring that for different arguments appearing in events, they should pay attention to context information highly relevant to the entity~\cite{zhou2021document} and target event~\cite{ebner2020multi}. Therefore
in this paper, we design a \textbf{S}pan-\textbf{T}rigger-based
\textbf{C}ontextual \textbf{P}ooling (STCP) module, which merges the information of non-argument clues for each argument-trigger pair based on the their contextual attention product from pre-trained model, enhancing the candidate argument representation with additional relevant context information. 

\begin{figure}[tbp]
    \centering
    \includegraphics[width=0.95\linewidth]{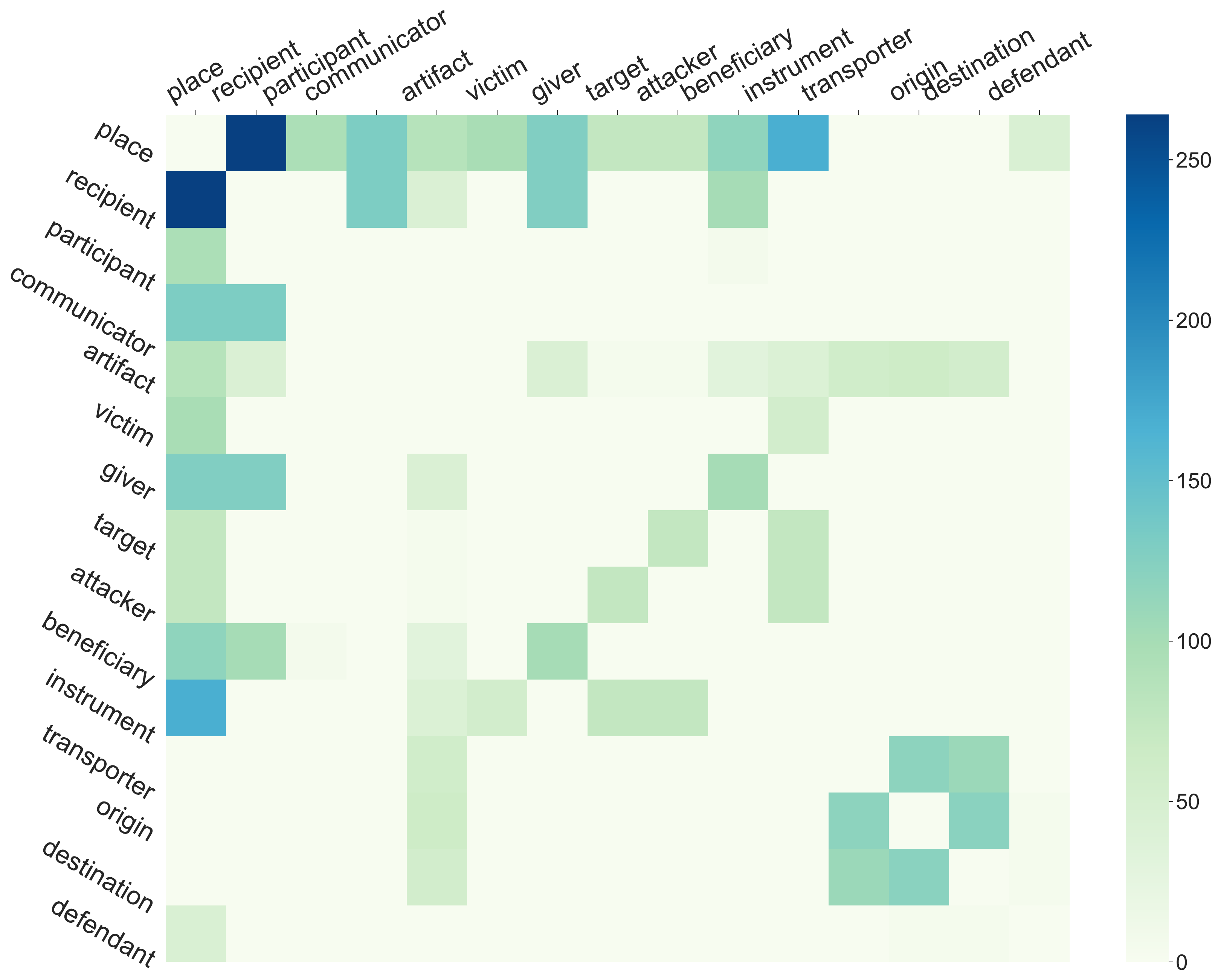}
    
    \caption{
Visualization of the co-occurrence frequency
 between 15 most frequent roles on
RAMS test set. we have reserved and set the co-occurrence number with itself to zero. The full ﬁgure
is included in Appendix~\ref{sec:appendixC}.}
    \label{fig:fig-2}
\end{figure}

Some argument roles have close semantic relevance that is beneficial for argument extraction. For example, in Figure~\ref{fig:fig1}, there is close semantic relevance between roles \textit{injurer} and \textit{victim}, which can provide significant information guidance for the argument extraction of these two roles in the target event \textit{Conflict and Attack}.
Moreover, many roles co-occur in multiple events~\cite{ebner2020multi,lietal2021document}, which may have close semantic relevance. 
Specifically, we count and visualize the frequency of co-occurrence between 15 most frequent roles in RAMS dataset in Figure~\ref{fig:fig-2}. For example,  roles \textit{attacker}, \textit{target} and \textit{instrument} co-occur frequently, demonstrating that they are more semantically relevant than other roles. In this paper, we propose a \textbf{R}ole-based \textbf{L}atent \textbf{I}nformation \textbf{G}uidance (RLIG) module, consisting of role-interactive encoding and role information fusion. Specifically, we design a role-interactive encoder with roles added into the input sequence, where role embeddings can not only learn latent semantic information of roles, but capture  semantic relevance among roles. 
The latent role embeddings are then merged into candidate arguments through pooling and concatenating operations, providing information guidance for document-level EAE.

In this paper, we propose an effective document-level EAE model named SCPRG (\textbf{S}pan-trigger-based
\textbf{C}ontextual \textbf{P}ooling and \textbf{R}ole-based
latent information \textbf{G}uidance) containing STCP module and RLIG module for the the aforementioned two problems respectively. 
Notably, these two modules leverage the well-learned attention weights from the pre-trained language model with no more than 1\% new parameters introduced and are easily applied to other event extraction models, which are compact and transplantable.
Moreover, we try to eliminate noise information by excluding argument-impossible spans. Our contributions are summarized as follows:

\begin{itemize} 
\item We propose a span-trigger-based contextual pooling module, which adaptively selects and aggregates the information of non-argument clues, enhancing the candidate argument representation with relevant context information.
\item We propose a role-based latent information guidance module, which provides latent role information guidance containing semantic relevance among roles.
\item Extensive experiments show that SCPRG outperforms previous
start-of-the-art models, with 1.13 F1 and 2.64 F1 improvements on public RAMS and WikiEvents~\cite{lietal2021document} datasets. We further analyse the attention weights and latent role representations, which shows the interpretability of our model\footnote{Our implementation is available at \url{https://github.com/LWL-cpu/SCPRG-master}}.

\end{itemize}

\begin{figure*} [htbp]
  \centering
  \includegraphics[width=1.05\linewidth]{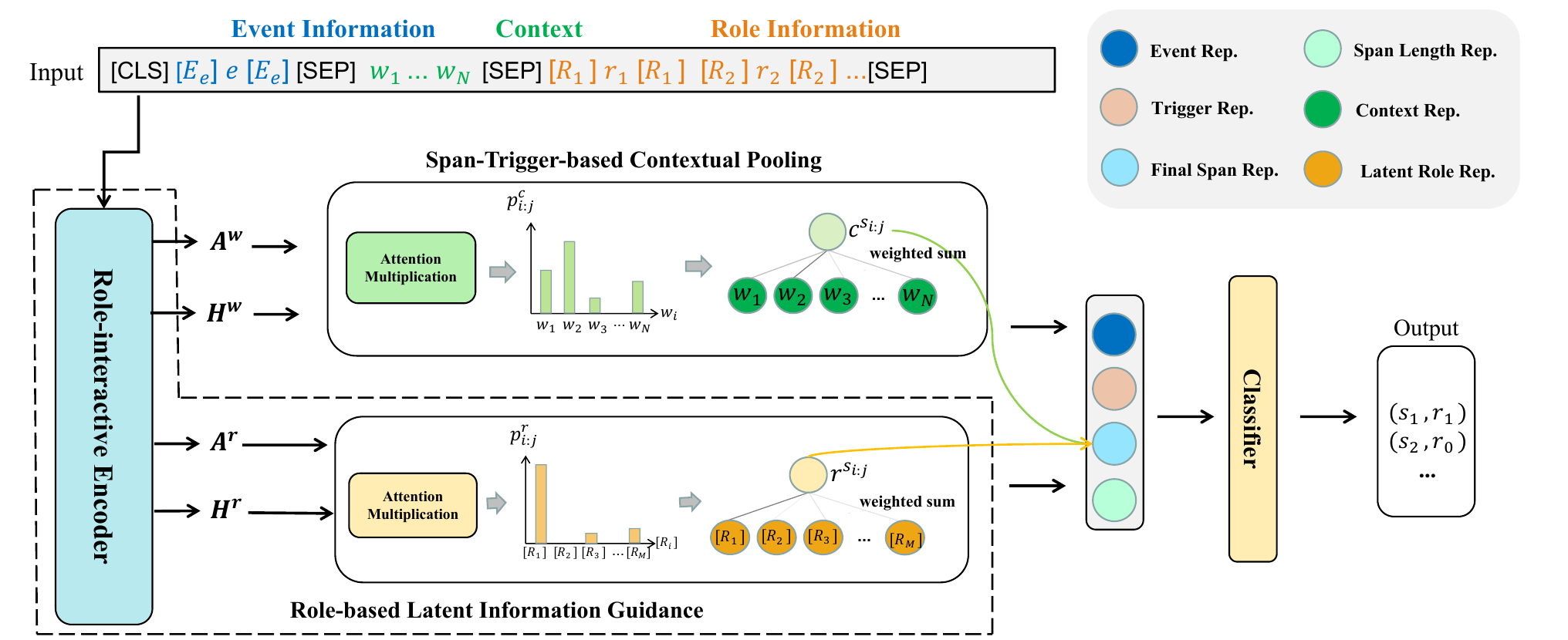} 
  \caption{The main architecture of SCPRG. The input sequence with roles is fed into the role-interactive encoder, with  context representations, role representations and attention heads as output. STCP adaptively fuses non-argument contextual
clues into a context vector based on the attention product between the trigger and arguments. RLIG constructs latent role embeddings through role-interactive encoding  and fuses them into a latent role vector by pooling operation.  The context vector and latent role vector are merged into the final span representation and the classiﬁcation module 
 predicts argument roles for all candidate spans.} 
  \label{fig:fig2} 
\end{figure*}

\section{Method}

We formulate document-level event argument extraction as a multi-class classification problem. Given a document $\mathcal{D}$ consisting of $N$ words, i.e. $\mathcal{D} = \{{w}_1, {w}_2, ..., {w}_{N}\}$, pre-defined event types set $\mathcal{E}$, the corresponding role set $\mathcal{R}_e$ and trigger $t \in \mathcal{D}$ for each event $e \in \mathcal{E}$, this task aims at predicting all $(r,s)$ pairs for each event in document $\mathcal{D}$, where $r \in \mathcal{R}_e$ is an argument role for event $e \in \mathcal{E}$ and $s \subseteq \mathcal{D}$ is a contiguous text span in $\mathcal{D}$. Following~\cite{ebner2020multi, xu2022two}, we extract event arguments for each event in a document independently and Figure~\ref{fig:fig2} shows the overall architecture of our
SCPRG.

\subsection{Role-interactive Encoder}
\label{role_encoder}
\textbf{Role Type Representation} \quad
In order to capture semantic relevance among roles, we add role type information into the input sequence and make interaction among context and roles  by multi-head attention, which obtains context and role representations in a shared knowledge space. 
Specifically, we construct latent embeddings of roles with different special tokens \footnote{In our implement, we utilize [unused] tokens for BERT~\cite{devlinetal2019bert} and add special tokens for RoBERTa~\cite{DBLP:journals/corr/abs-1907-11692}.} in the pre-trained model, where each role type has a specific latent representation. On account that role names also contain valuable semantic information~\cite{wangetal2022query}, we wrap role names with special role type tokens and take the embedding of the start special toke as the
role embedding. Taking the role \textit{$Place$} as an example, we ﬁnally represent it as \textit{$[R_0] \ Place \ [R_0]$}, where \textit{$[R_0]$} is the special role type token of \textit{$Place$}.

\noindent \textbf{Role-interactive Encoding} \quad
For the input document $\mathcal{D} = \{{w}_1, {w}_2, ..., {w}_{N}\}$, the target event $e$ and the corresponding role set $\mathcal{R}_e = \{{r}_1, {r}_2, {r}_3, ...\}$, we concatenate them into a sequence as follows:

 $ S \ =  \ \mathrm{[CLS]} \  [E_e] \ {e}  \ [E_e] \ \mathrm{[SEP]} \  w_1 \  ... \  w_N \ \mathrm{[SEP]}$ 
 
\quad \quad \quad \textsc{ $[R_1] \ {r_1} \   [R_1] \  [R_2] \  {r_2} \  [R_2] \  ... \  \mathrm{[SEP]}$},

\noindent where \textsc{$[E_e]$} is the special event token of event $e$. \textit{$[R_1]$} and \textit{$[R_2]$} are the special role type tokens of $r_1$ and $r_2$. We use the last $\mathrm{[SEP]}$ to represent none category. Next, we leverage the pre-trained language model as an encoder to obtain the embedding of each token as follows:

\begin{equation}
\setlength\abovedisplayskip{3pt plus 3pt minus 7pt}
\setlength\belowdisplayskip{3pt plus 3pt minus 7pt}
    \mathbf{H}^s = \mathrm{Encoder} (\textsc{$S$}).
\end{equation}
Then we can obtain event representation $\mathbf{H}^e \in \mathbb{R}^{1\times d}$ of the start \textsc{$[E_e]$}, context representation $\mathbf{H}^w \in \mathbb{R}^{l_w\times d}$, and role representation $\mathbf{H}^r \in \mathbb{R}^{l_r\times d}$ respectively from $\mathbf{H}^s$, where $l_w$ is the length of word pieces list and $l_r$ is the length of role list. 
For input sequences longer than 512, we leverage a dynamic
window to encode the whole sequence and average the overlapping token embeddings of different windows to obtain the final representation.

Significantly, through role-interactive encoding, the role embeddings can capture semantic relevance and
adapt to the target event and context, which better guides the argument extraction.

\subsection{Span-Trigger-based Contextual Pooling}

\textbf{Argument-impossible Spans Exclusion} \quad
In order to eliminate noise information of useless spans, we reduce the number of candidate spans by excluding some argument-impossible spans, e.g. spans with comma in the middle. With such improvement, we reduce a quarter of candidate spans on average and make our model pay attention to candidate spans with useful information. 

\noindent \textbf{Span-Trigger-based Contextual Pooling} \quad
For a candidate span ranging from $w_i$ to $w_j$, most previous span-based methods\cite{zhang2020two, xu2022two} represent it through the average pooling of the hidden state of tokens within this span:
     $\frac1{j-i+1}\sum_{k=i}^j\mathbf{h}_k^w,$
where $\mathbf{h}_k^w$ is the $k^{\mathrm{th}}$ token embedding from $\mathbf{H}^w$.

However, average pooling representation ignores the significant clue information of other non-argument words. Although self-attention mechanism of the pre-trained encoder can model token-level interaction, such global interaction is specific to the event and candidate arguments.  Therefore, we propose to select and fuse useful contextual information highly related to each tuple consisting of a candidate span and the event trigger word, i.e. $(s_{i:j}, t)$. We directly utilize the attention heads of pre-trained transformer-based encoder for span-trigger-based contextual pooling, which transfers the well-learned dependencies from the pre-trained language model without learning new
attention layers from scratch~\cite{zhou2021document}.

Specifically, we use the token-level attention
heads $\textbf{A}^{w} \in \mathbb{R}^{H\times l_w\times l_w}$ of context from the last transformer layer in the pre-trained language model. Then we can obtain the context attention $\textbf{A}^{C}_{i:j} \in \mathbb{R}^{l_w}$ of each candidate span ranging from $w_i$ to $w_j$ with average pooling:

\begin{equation}
\label{eq3}
\setlength\abovedisplayskip{3pt plus 3pt minus 7pt}
\setlength\belowdisplayskip{3pt plus 3pt minus 7pt}
     \textbf{A}^{C}_{i:j} = \frac1{H(j-i+1)}\sum_{h=1}^H\sum_{m=i}^j\mathbf{A}^{w}_{h,m}.
\end{equation}

Then for span-trigger pair $(s_{i:j}, t)$, we obtain the contextual clue information  $\textbf{c}^{s_{i:j}} \in \mathbb{R}^{d}$ that are important to candidate span by multiplying the attentions followed by normalization:

\begin{equation}
\label{eq4}
\setlength\abovedisplayskip{3pt plus 3pt minus 7pt}
\setlength\belowdisplayskip{3pt plus 3pt minus 7pt}
    \begin{array}{c}
    \textbf{p}^{c}_{i:j}=softmax(\textbf{A}^{C}_{i:j}\cdot \textbf{A}_{t}^{C}\;)\vspace{1.3ex}, \\
    \textbf{c}^{s_{i:j}}={\mathbf{H}^w}^T \ \textbf{p}^{c}_{i:j},
    \end{array}
\end{equation}%
where $\textbf{A}^{C}_{t} \in \mathbb{R}^{l_w}$ is the contextual attention of trigger $t$ and $\textbf{p}^{c}_{i:j} \in \mathbb{R}^{l_w}$ is the computed attention weight vector for context. $T$ is the transpose symbol.
\subsection{Role-based Latent Information Guidance}

RLIG module constructs latent role embeddings through role-interactive encoding in Sec.~\ref{role_encoder} and performs role information fusion through pooling operation, which provides valuable latent role information guidance.

\noindent \textbf{Role Information Fusion} \quad
In order to make each candidate argument get the useful information guidance of roles, we modify our span-trigger-based contextual pooling method to select role information adaptively. 
We get the latent role information $\textbf{r}^{s_{i:j}} \in \mathbb{R}^{d}$ for ${s_{i:j}}$
through contextual pooling, by modifying the operation in Eq.~\ref{eq3} and Eq.~\ref{eq4}:

\begin{equation}
\label{eq7}
\setlength\abovedisplayskip{3pt plus 3pt minus 7pt}
\setlength\belowdisplayskip{3pt plus 3pt minus 7pt}
    \begin{array}{c}

    \textbf{A}^{R}_{i:j} = \frac1{H(j-i+1)}\sum_{h=1}^H\sum_{m=i}^j\mathbf{A}^{r}_{h,m}\vspace{1.3ex},\\
    \textbf{p}^{r}_{i:j}=softmax(\textbf{A}^{R}_{i:j}\cdot \textbf{A}_{t}^{R}\;)\vspace{1.3ex}, \\
    \textbf{r}^{s_{i:j}}={\mathbf{H}^r}^T \ \textbf{p}^{r}_{i:j},
    \end{array}
\end{equation}%
where $\textbf{A}^{r} \in \mathbb{R}^{H\times l_w\times l_r}$ are attention heads of roles from the
last transformer layer in the pre-trained language
model. $\textbf{A}^{R}_{i:j} \in \mathbb{R}^{l_r}$ is the role attention for each candidate span and  $\textbf{A}^{R}_{t} \in \mathbb{R}^{l_r}$ is the role attention of trigger $t$. $\textbf{p}^{r}_{i:j} \in \mathbb{R}^{l_r}$ is the computed attention weight vector for roles.

For a candidate span ${s_{i:j}}$, we fuse the average pooling representation, contextual clue information $\textbf{c}^{s_{i:j}}$ and latent role information $\textbf{r}^{s_{i:j}}$ as follows:
\begin{equation} 
\setlength\abovedisplayskip{3pt plus 3pt minus 7pt}
\setlength\belowdisplayskip{3pt plus 3pt minus 7pt}
\textbf{s}_{i:j}=tanh(\mathbf{W}_{1}[\frac1{j-i+1}\sum_{k=i}^j\mathbf{h}_k^w; \textbf{c}^{s_{i:j}};\textbf{r}^{s_{i:j}}]),
\end{equation}
where $\mathbf{W}_{1} \in \mathbb{R}^{3d \times d}$ is learnable parameter.

\begin{table*}[]
\centering
\setlength{\tabcolsep}{1.0mm}{
\centering
\begin{tabular}{llccccc}
\hline
\textbf{Dataset} & \textbf{Split} & \multicolumn{1}{l}{\textbf{\# Doc.}} & \multicolumn{1}{l}{\textbf{\# Event}} & \multicolumn{1}{l}{\textbf{\# Argument}}
& \multicolumn{1}{l}{\textbf{\# Event Types}}
& \multicolumn{1}{l}{\textbf{\# Role Types}}\\ \hline
\multicolumn{1}{c}{} & Train & 3,194 & 7,329 & 17,026 & 139 & 65\\
RAMS                 & Dev   & 399   & 924   & 2,188  & 131 & 62\\
\multicolumn{1}{c}{} & Test  & 400   & 871   & 2,023  & - & - \\ \hline
                     & Train & 206   & 3,241 & 4,542  & 49 & 57\\
WikiEvents           & Dev   & 20    & 345   & 428   & 35 & 32 \\
                     & Test  & 20    & 365   & 566   &34 &44 \\ \hline
\end{tabular}
}

\caption{Detailed statistics of two datasets.}
\label{tab:table1}
\end{table*}

\subsection{Classiﬁcation Module}
\noindent \textbf{Boundary Loss} \quad
Since we extract arguments in span level, whose
boundary may be ambiguous, we construct start and end representation with fully connected neural networks to enhance the representation of candidate spans: $\mathbf{H}^{start}=\mathbf{W}^{start}\mathbf{H}^s, \mathbf{H}^{end}=\mathbf{W}^{end}\mathbf{H}^s$,
where $\mathbf{H}^s$ is the hidden representation of input sequence \textsc{S}. 
On this basis, we enhance the start and end representation by integrating context and role information with span-trigger-based contextual pooling as follows:

\begin{equation}
\setlength\abovedisplayskip{7pt plus 8pt minus 11pt} 
\setlength\belowdisplayskip{9pt plus 10pt minus 12pt}
\label{eq9}
    \begin{array}{c}
    \textbf{z}^{start}_{i:j}={\mathbf{H}^{start}}^T \ \textbf{p}_{i:j}\vspace{1.0ex}, \\
    \textbf{z}^{end}_{i:j}={\mathbf{H}^{end}}^T \ \textbf{p}_{i:j}\vspace{1.0ex}, \\
    \mathbf{h}^{start}_{i:j}=tanh(\mathbf{W}_{2}[\mathbf{h}_i^{start}; \textbf{z}^{start}_{i:j}])\vspace{1.0ex},\\
    \mathbf{h}^{end}_{i:j}=tanh(\mathbf{W}_{3}[\mathbf{h}_j^{end}; \textbf{z}^{end}_{i:j}]), \vspace{1.0ex}
    \end{array}
\end{equation}%
where $\mathbf{h}^{start}_i$ and $\mathbf{h}^{end}_j$ are the 
$i^{\mathrm{th}}$ and $j^{\mathrm{th}}$
vector of  $\mathbf{H}^{start}$ and $\mathbf{H}^{end}$.  $\textbf{p}_{i:j}$ is the computed attention vector for both context and roles which is calculated similarly to Eq.~\ref{eq4} or Eq.~\ref{eq7} and
$\mathbf{W}_2, \mathbf{W}_3 \in \mathbb{R}^{2d\times d}$ are learnable parameters.
Then we obtain the final representation $\widetilde{\mathbf s}_{i:j}$ for a candidate span as follows: $\widetilde{\mathbf s}_{i:j}=\mathbf{W}^{s}[\mathbf{h}^{start}_{i:j}; \textbf{s}_{i:j}; \mathbf{h}^{end}_{i:j}],$
where $\mathbf{W}^{s} \in \mathbb{R}^{3d\times d}$ is the learnable model parameter. 

Finally, the boundary loss is deﬁned to detect the start and end position following~\cite{xu2022two}:
\begin{equation}
\label{eq12}
 \begin{array}{c}
\mathcal{L}_b=-\sum_{i=1}^{\vert \mathcal{D}\vert}\lbrack y_i^s\log P_i^s+(1-y_i^s)\log(1-P_i^s) \vspace{1.3ex} \\
+y_i^e\log P_i^e+(1-y_i^e)\log(1-P_i^e)\rbrack,
\end{array}
\end{equation}%
where $y_i^s$ and $y_i^e$ denote golden labels and $P_i^s = \mathrm{sigmoid}(\mathbf{W}_4  \mathbf{h}^{start}_i)$ and $P_i^e = \mathrm{sigmoid}(\mathbf{W}_5  \mathbf{h}^{end}_i)$ are the probabilities of the word
$w_i$ predicted to be the ﬁrst or last word of a golden argument span.

\noindent \textbf{Classification Loss} \quad
For a candidate span $s_{i:j}$ in event $e$, we concatenate the span representation  $\widetilde{\mathbf s}_{i:j}$, trigger representation $\mathbf{h}_t$, their absolute difference $|\mathbf{h}_t-\widetilde{\mathbf s}_{i:j}|$, element-wise multiplication $\mathbf{h}_t\odot\widetilde{\mathbf s}_{i:j}$, event type embedding $\mathbf{H}^e$ and span length embedding $\mathbf{E}_{len}$ and get the prediction $P(r_{i:j})$ of the candidate span $s_{i:j}$ via a feed-forward network:

\begin{equation}
\label{eq13}
\setlength\abovedisplayskip{3pt plus 3pt minus 7pt}
\setlength\belowdisplayskip{3pt plus 3pt minus 7pt}
\mathbf{I}_{i:j}=[\widetilde{\mathbf s}_{i:j}; \mathbf{h}_t; |\mathbf{h}_t-\widetilde{\mathbf s}_{i:j}|; \mathbf{h}_t\odot\widetilde{\mathbf s}_{i:j}; \mathbf{H}^e; \mathbf{E}_{len} ],
\end{equation}%

\begin{equation}
\label{eq14}
\setlength\abovedisplayskip{3pt plus 3pt minus 7pt}
\setlength\belowdisplayskip{3pt plus 3pt minus 7pt}
P(r_{i:j}) = \mathrm{FFN}(\mathbf{I}_{i:j}).
\end{equation}%

Considering most candidate arguments are negative samples and the imbalanced role distribution, we adopt focal loss~\cite{8237586} to make the training process focus more on useful positive samples, where $\alpha$ and $\gamma$ are hyperparameters.

\begin{equation}
\label{eq15}
\setlength\abovedisplayskip{3pt plus 3pt minus 7pt}
\setlength\belowdisplayskip{3pt plus 3pt minus 7pt}
 \begin{array}{c}
\mathcal{L}_c=-\sum_{i=1}^{\vert \mathcal{D}\vert}\sum_{j=1}^{\vert \mathcal{D}\vert}\alpha{{\lbrack1-P(r_{i:j}=y_{i:j}))}\rbrack}^\gamma \vspace{1.3ex}\\ 
\cdot\log P(r_{i:j}=y_{i:j}).
\end{array}
\end{equation}%

Finally, we have the train loss consisting of $\mathcal{L}_c$ and $\mathcal{L}_b$ with hyperparameter $\lambda$:
\begin{equation}
\label{eq16}
\mathcal{L}=\mathcal{L}_c+\lambda\mathcal{L}_b.
\end{equation}%

\section{Experiments}
\subsection{Experimental Setup}
\noindent \textbf{Datasets and Metrics} \quad
We evaluate the proposed model on two large-scale public document-level
EAE datasets, RAMSv1.0~\cite{ebner2020multi} and WikiEvents~\cite{lietal2021document} following the ofﬁcial
 train/dev/test split, whose detailed data statistic are shown in Table~\ref{tab:table1}. Following~\cite{xu2022two}, we report the Span F1 and Head F1 on dev and test sets for RAMS dataset. Span
F1 requires the predicted argument spans to fully
match the golden ones, while Head F1 evaluates solely on the head word\footnote{The head word of a span is deﬁned as the word that has the smallest arc distance
to the root in the dependency tree.} of the argument span. Additionally, for WikiEvents dataset, we report the Head F1 and Coref F1 scores on test set for argument identification task (Arg IF) and argument classification (Arg CF) task respectively following~\cite{lietal2021document}. The Coref F1 evaluates the coreference between extracted arguments and golden arguments as used by~\cite{jigrishman2008refining} and the model achieves Coref F1 if extracted arguments are coreferential with golden arguments.

\noindent \textbf{Baselines} \quad
We compare different categories of document-level EAE models which mainly consist of tagging-based methods such as \textbf{BERT-CRF}~\cite{DBLP:journals/corr/abs-1904-05255}, $\textbf{BERT-CRF}_{\text{TCD}}$~\cite{ebner2020multi}, span-based methods like \textbf{Two-Step}~\cite{zhang2020two}, $\textbf{Two-Step}_{\text{TCD}}$~\cite{ebner2020multi}, \textbf{TSAR}~\cite{xu2022two}, and other generation-based methods such as \textbf{FEAE}~\cite{weietal2021trigger}, \textbf{BERT-QA}~\cite{ducardie2020event}, \textbf{BART-Gen}~\cite{lietal2021document},
$\textbf{EA}^2$\textbf{E}~\cite{zengetal2022ea2e}. Moreover, we use ${\text{BERT}}_{{\text{base}}}$~\cite{devlinetal2019bert} and ${\text{RoBERTa}}_{{\text{large}}}$~\cite{DBLP:journals/corr/abs-1907-11692} as the pre-trained transformer-based encoder.

\noindent \textbf{Hyperparameters Setting} \quad
We set the dropout rate to 0.1, batch size to 8, and
train our SCPRG using Adam~\cite{Kingma2014AdamAM} as
optimizer with 3e-5 learning rate. The hidden dimension $d$  is 768 for ${\text{SCPRG}}_{{\text{base}}}$ and 1024 for ${\text{SCPRG}}_{{\text{large}}}$.  In order to mitigate imbalanced role distribution problem, we set the weight ratio $\alpha$ of empty class and other classes to 10:1. We set hyperparameters $\gamma$ to 2 and boundary loss weight $\lambda$ to 0.1 for both two datasets. We train SCPRG
for 50 epochs for RAMS dataset and 100 epochs
for WikiEvents dataset.
\subsection{Main Results}
Table~\ref{tab:table2} shows the experimental results on both dev and test set in RAMS dataset. Compared with previous tagging-based and span-based methods like BERT-CRF and Two-Step, our SCPRG equipped with ${\text{BERT}}_{{\text{base}}}$ yields an improvement of \textbf{+8.46/+9.64} $ \sim $  \textbf{+6.36/+7.14} Span F1 and \textbf{+7.68/+9.00} $ \sim $  \textbf{+5.38/+6.40} Head F1 on dev/test set, showing that our SCPRG framework has superiority in excluding impossible candidate spans and solving the imbalance of data distribution problem.
Significantly, SCPRG with ${\text{RoBERTa}}_{{\text{large}}}$ also outperforms previous state-of-the-art models ${\text{BART-Gen}}_{{\text{large}}}$\footnote{${\text{BART-Gen}}_{{\text{large}}}$ is based on ${\text{BART}}_{{\text{large}}}$~\cite{DBLP:journals/corr/abs-1910-13461} which is pre-trained on the same corpus.
} (\textbf{+3.68/+2.34} Span/Head F1 on test set) and ${\text{TSAR}}_{{\text{large}}}$ (\textbf{+1.14/+1.13} Span/Head F1 on test set). These results demonstrate the superior extraction ability of our model, benefiting from the effect of contextual clue information and latent role representation with semantic relevance.

Moreover, we further validate our SCPRG on WikiEvents and achieve new state-of-the-art performance in both tasks with base and large pre-trained models,
which can be viewed in Table~\ref{tab:table3}. Our SCPRG outperforms previous competitive methods like TSAR and $\text{EA}^2\text{E}$.  Compared with ${\text{TSAR}}_{{\text{large}}}$, our SCPRG improves up to \textbf{+0.64/+0.58} Head/Coref F1 for argument identiﬁcation and \textbf{+1.22/+1.29} Head/Coref F1 for argument classiﬁcation on the test set. Besides, SCPRG also outperforms recent competitive generation-based method ${\text{EA}^2\text{E}}_{{\text{large}}}$ in argument identiﬁcation (\textbf{+2.64/+0.33} Head/Coref F1) and argument classiﬁcation (\textbf{+2.31/+0.38} Head/Coref F1) tasks. These experimental improvements demonstrate the great advantage of our framework fused with argument-event specific context information and the helpful guidance of latent role information.
\begin{table}[tbp]
\setlength{\tabcolsep}{0.6mm}{
\begin{tabular}{lcccc}

\hline
\multirow{2}{*}{\textbf{Method}}  &
  \multicolumn{2}{c}{\textbf{Dev}} &
  \multicolumn{2}{c}{\textbf{Test}} \\ \cline{2-5} 
 &
  \multicolumn{1}{c}{\footnotesize{Span F1}} &
  \multicolumn{1}{c}{\footnotesize{Head F1}} &
  \multicolumn{1}{c}{\footnotesize{Span F1}} &
  \multicolumn{1}{c}{\footnotesize{Head F1}} \\ \hline
BERT-CRF     & 38.1           & 45.7           & 39.3           & 47.1           \\
${\text{BERT-CRF}}_{{\text{TCD}}}$ & 39.2           & 46.7           & 40.5           & 48.0           \\
Two-Step     & 38.9           & 46.4           & 40.1           & 47.7           \\
${\text{Two-Step}}_{{\text{TCD}}}$ & 40.3           & 48.0           & 41.8           & 49.7           \\
${\text{TSAR}}_{{\text{base}}}$         & 45.23          & 51.70          & 48.06          & 55.04          \\
FEAE              & -              & -              & 47.40          & -              \\

${\text{SCPRG}}_{{\text{base}}}$ (Ours)        & \textbf{46.56} & \textbf{53.38} & \textbf{48.94} & \textbf{56.10} \\ \hline
${\text{BART-Gen}}_{{\text{large}}}$    & -              & -              & 48.64          & 57.32          \\
${\text{TSAR}}_{{\text{large}}}$        & 49.23          & 56.76          & 51.18          & 58.53          \\
${\text{SCPRG}}_{{\text{large}}}$ (Ours)        & \textbf{50.53} & \textbf{57.66} & \textbf{52.32} & \textbf{59.66} \\ \hline
\end{tabular}
}
\caption{Main results of RAMS.}
\label{tab:table2}
\end{table}


\subsection{Ablation Study}
To better illustrate the capabilities of
our components, we conduct ablation study on RAMS dataset as shown in Table~\ref{tab:table4}.  We also provide ablation study results on WikiEvents datasets in Appendix~\ref{sec:appendixB}.

First, when we remove span-trigger-based contextual pooling (STCP) module, both Span F1 and Head F1 score of ${\text{SCPRG}}_{{\text{base}}}$/ ${\text{SCPRG}}_{{\text{large}}}$ drop by \textbf{1.61/1.43} and
\textbf{1.42/2.09} on test set, which indicates that our STCP plays a vital role in capturing the clue information of non-argument context that is crucial for document-level EAE. 

\begin{table}[tbp]
\setlength{\tabcolsep}{0.55mm}{
\begin{tabular}{lcccc}
\hline
\multirow{2}{*}{\textbf{Method}} & \multicolumn{2}{c}{\textbf{Arg IF}} & \multicolumn{2}{c}{\textbf{Arg CF}} \\ \cline{2-5} 
            & \footnotesize{Head F1} & \footnotesize{Coref F1} & \footnotesize{Head F1} & \footnotesize{Coref F1} \\ \hline
BERT-CRF    & 69.83            & 72.24             & 54.48            & 56.72             \\
BERT-QA     & 61.05            & 64.59             & 56.16            & 59.36             \\
BERT-QA-Doc & 39.15            & 51.25             & 34.77            & 45.96             \\
${\text{TSAR}}_{{\text{base}}}$   & 75.52            & 73.17             & 68.11            & 66.31             \\
${\text{SCPRG}}_{{\text{base}}}$ (Ours)  & \textbf{76.13}   & \textbf{74.90}    & \textbf{68.91}   & \textbf{68.33}    \\ \hline
${\text{TSAR}}_{{\text{large}}}$  & 76.62            & 75.52             & 69.70            & 68.79  
\\
${\text{BART-Gen}}_{{\text{large}}}$    & 71.75            & 72.29             & 64.57            & 65.11             \\

${\text{EA}^2\text{E}}_{{\text{large}}}$    & 74.62            & 75.77             & 68.61            & 69.70             \\
${\text{SCPRG}}_{{\text{large}}}$ (Ours)  & \textbf{77.26}   & \textbf{76.10}    & \textbf{70.92}   & \textbf{70.08}    \\ \hline
\end{tabular}
}
\caption{Main results of WikiEvents.}
\label{tab:table3}
\end{table}

Additionally, when removing role-based latent information guidance (RLIG) module\footnote{We also remove the corresponding role tokens added in the input sequence.}, the performance of ${\text{SCPRG}}_{{\text{base}}}$/ ${\text{SCPRG}}_{{\text{large}}}$ drops sharply by \textbf{1.03/1.04} Span F1 and \textbf{1.58/1.2} Head F1 on
RAMS test set. It suggests that our RLIG module effectively guides argument extraction with meaningful latent role representations containing semantic relevance among roles. 
When removing both STCP and RLIG module, the performance decay exceeds that when removing a single module, which explains that our two modules can work together to improve the performance. 

Moreover, when removing argument-impossible spans exclusion (ASE) operation, both ${\text{SCPRG}}_{{\text{base}}}$ and ${\text{SCPRG}}_{{\text{large}}}$ have
a performance decay, which indicates
that excluding argument-impossible candidate spans  eliminates noise information and contributes to argument extraction. Focal Loss helps to balance the representation of positive and negative samples, facilitating smooth convergence of the model during training. However, it does not contribute to improving the performance of the model.

\begin{figure}[tbp]
    \centering
    \includegraphics[width=1.0\linewidth]{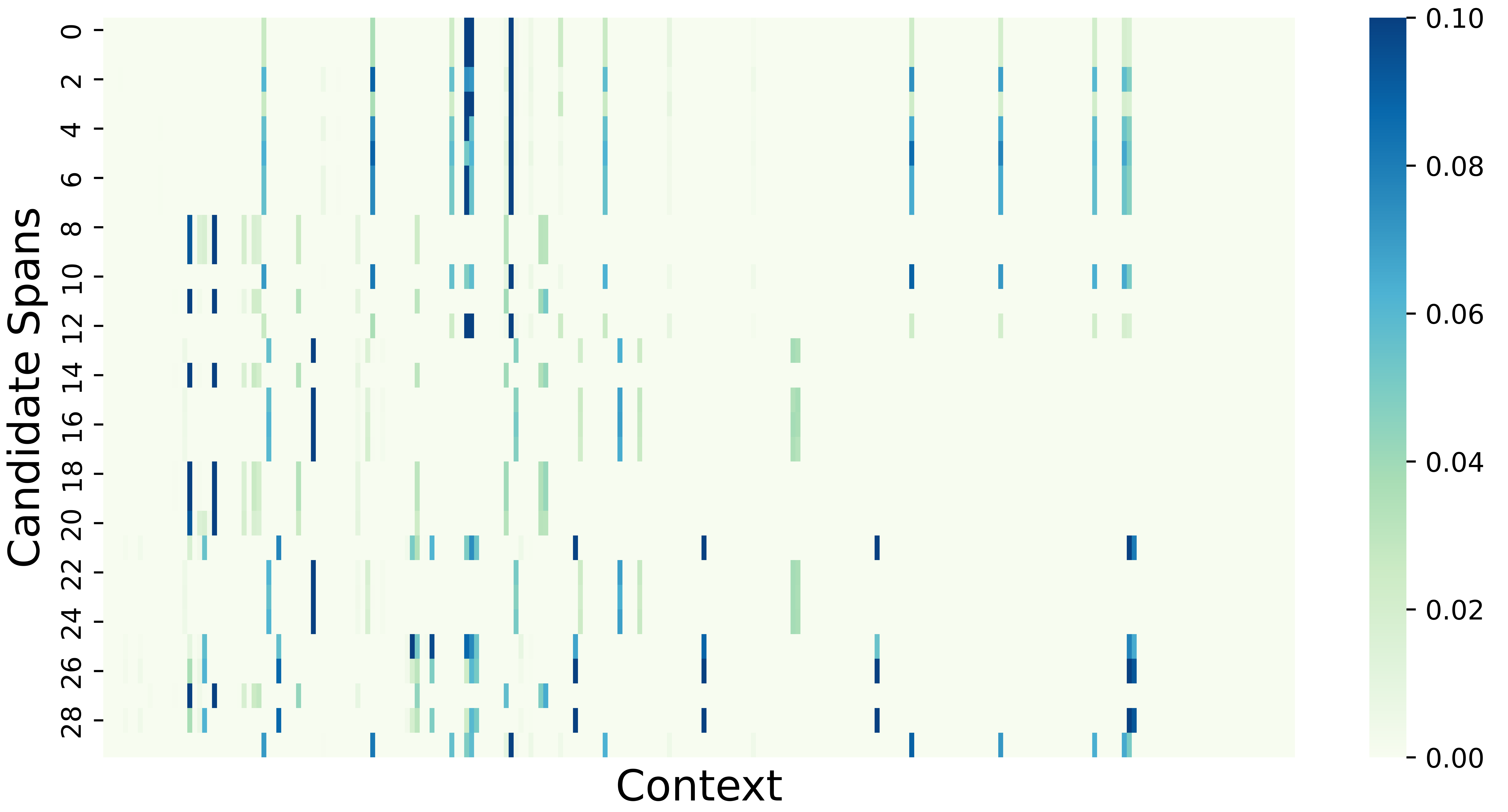}
    
    \caption{Visualization on attention weights to the context based on different candidate spans in an event.}
    \label{fig:fig5}
\end{figure}

\begin{figure*} [htbp]
  \centering
  \includegraphics[width=1.0\linewidth]{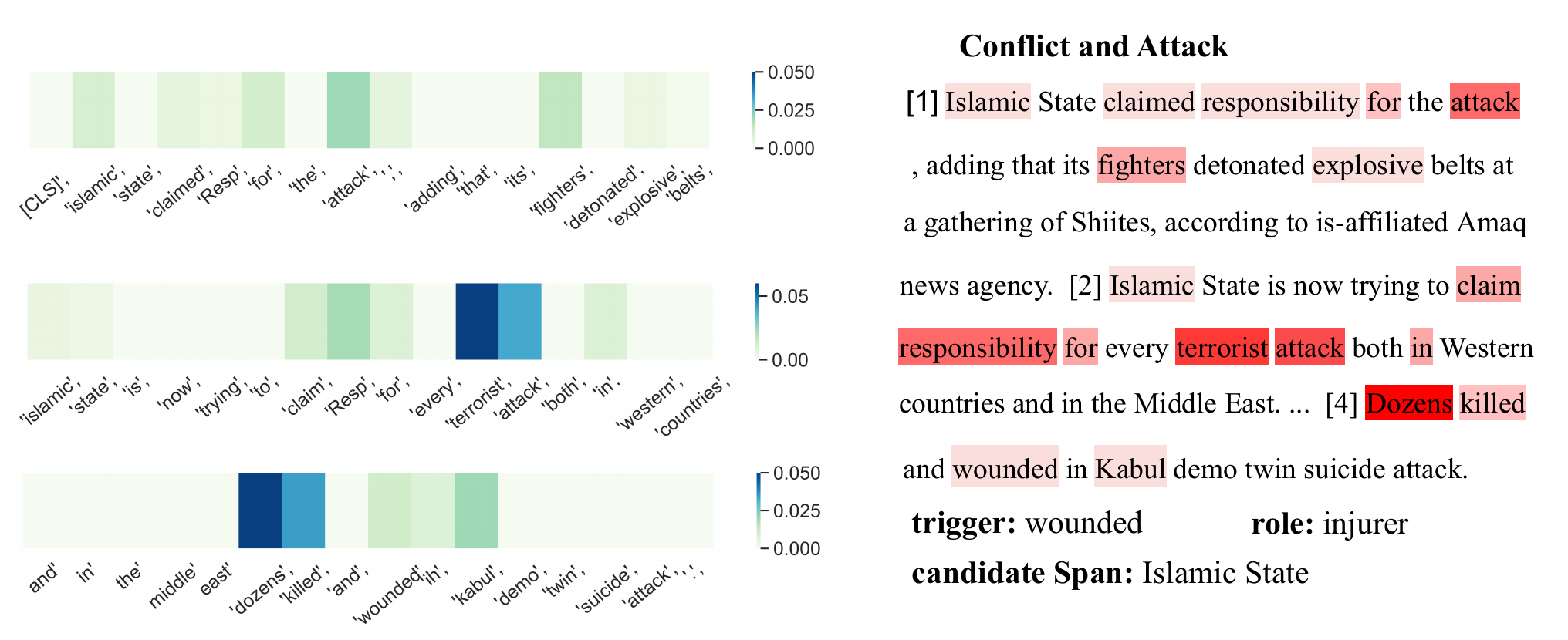} 
  \caption{Context weights of an example from RAMS. We visualize the weight of context tokens based on the span-trigger pair 
(\textit{Islamic State,
wounded}). We use different shades of color to represent attention weights. } 
  \label{fig:fig3} 
\end{figure*} 
\subsection{Analysis of Context Attention Weights}

To assess the effectiveness of STCP in capturing useful contextual information for candidate arguments, we visualize the contextual weights $\textbf{p}^{c}_{i:j}$ in Eq.~\ref{eq4} of an example of Figure~\ref{fig:fig1}. As shown in Figure~\ref{fig:fig3}, our
STCP gives high weights to non-argument words such as
\textit{attack}, \textit{responsibility} and \textit{terrorist attack}, which are most relevant to the span-trigger pair  (\textit{Islamic State,
wounded}). 
Interestingly, our STCP also gives relatively high attention weights to words in other arguments like \textit{explosive}, \textit{Dozens} and \textit{Kabul}, which means that these argument words provide important information for the role prediction of \textit{Islamic State}. The visualization demonstrates that our STCP can not only
capture the non-argument clue information that is related to candidate spans, but model the information interaction among related arguments in an event.

Additionally, we also explore the attention weights based on different span-trigger pairs in an event. In Figure~\ref{fig:fig5}, we randomly select 30 candidate spans in an event and draw the heat map based on their attention weights to the context. The heat map shows that different candidate arguments focus on different context information, indicating that our STCP can adaptively select contextual information according to candidate argument spans.

\subsection{Analysis of Role Information Guidance}

\begin{figure}[tbp]
    \centering
    \subfloat[\label{fig:a}]{
    \includegraphics[width=0.47\linewidth]{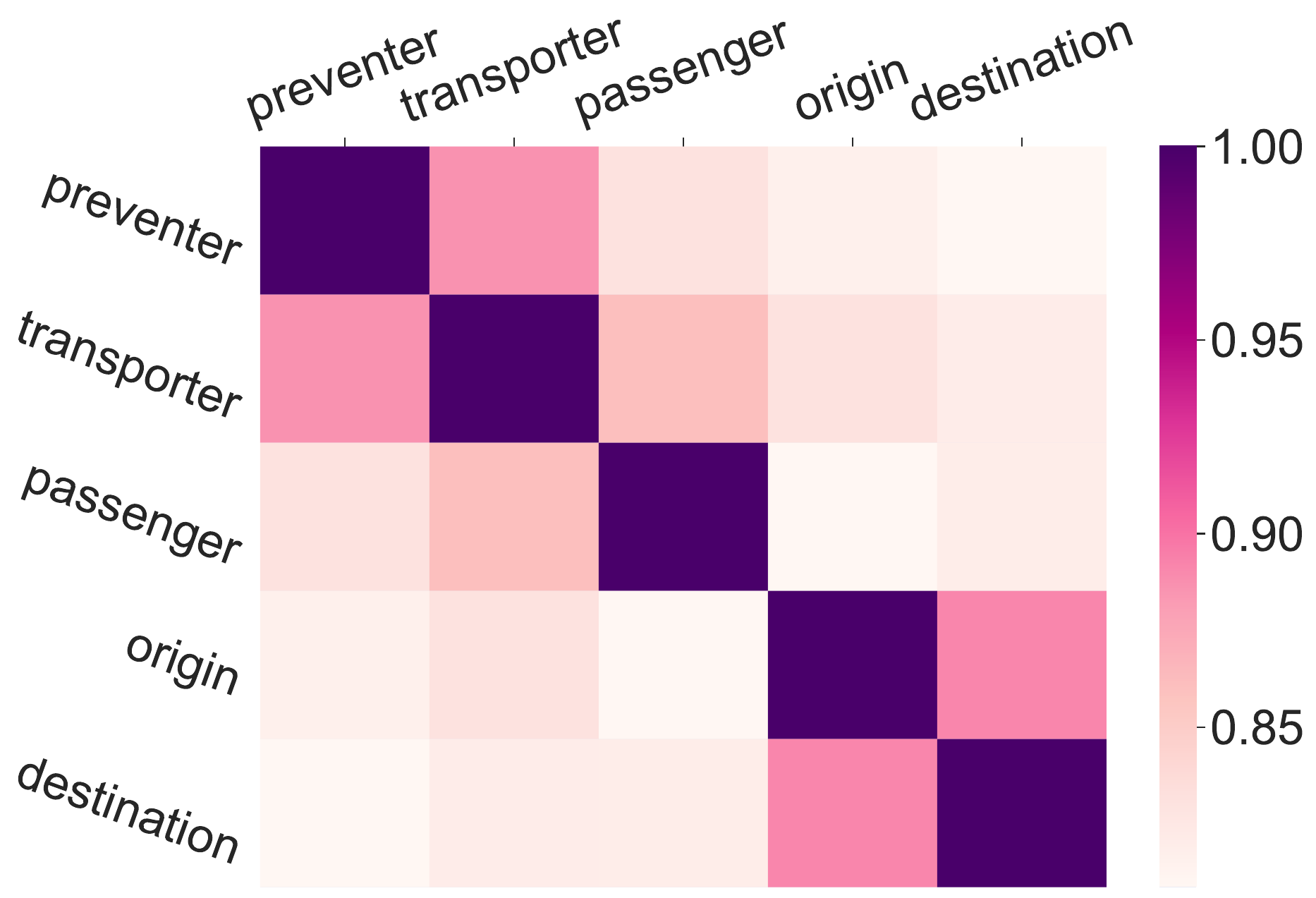}
    }
    \subfloat[\label{fig:b}]{
    \includegraphics[width=0.47\linewidth]{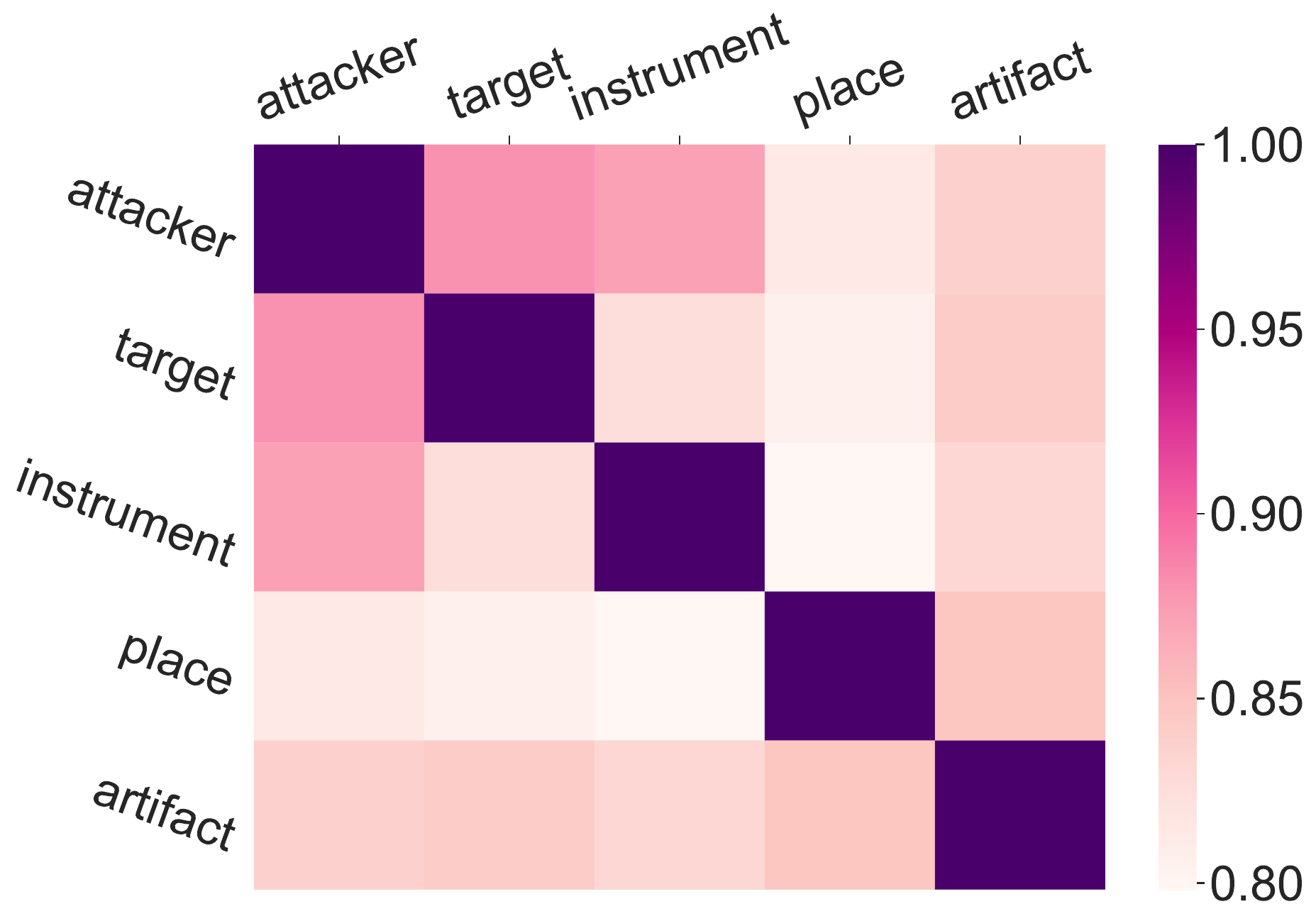}
    }
    \caption{The visualization of cosine similarity between role representations from two examples in RAMS dataset.}
    \label{fig:fig8}
\end{figure}

To verify that our model can capture semantic relevance among roles, we visualize the cosine similarity between latent role representations from two events in RAMS dataset in Fig~\ref{fig:fig8}. As the figure shows, roles \textit{origin} and \textit{destination}, \textit{attacker} and \textit{target} have similar representations, which agrees with their semantics, demonstrating that our model can  capture the semantic relevance among roles.

\begin{figure}[tbp]
    \centering
    \subfloat[\label{fig:c}Without latent role guidance.]{
    \includegraphics[width=0.5\linewidth]{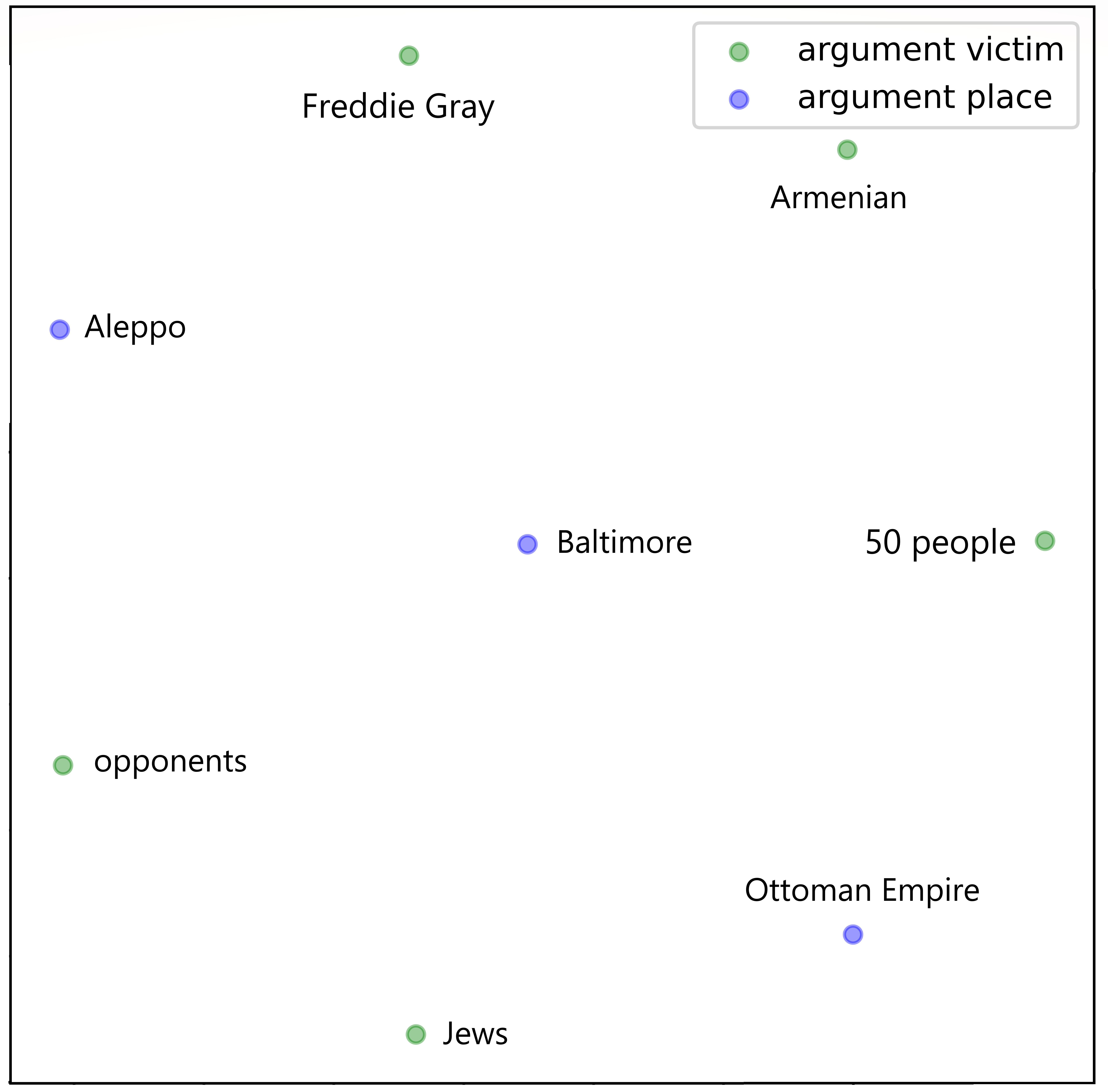}
    }
    \subfloat[\label{fig:d}With latent role guidance.]{
    \includegraphics[width=0.5\linewidth]{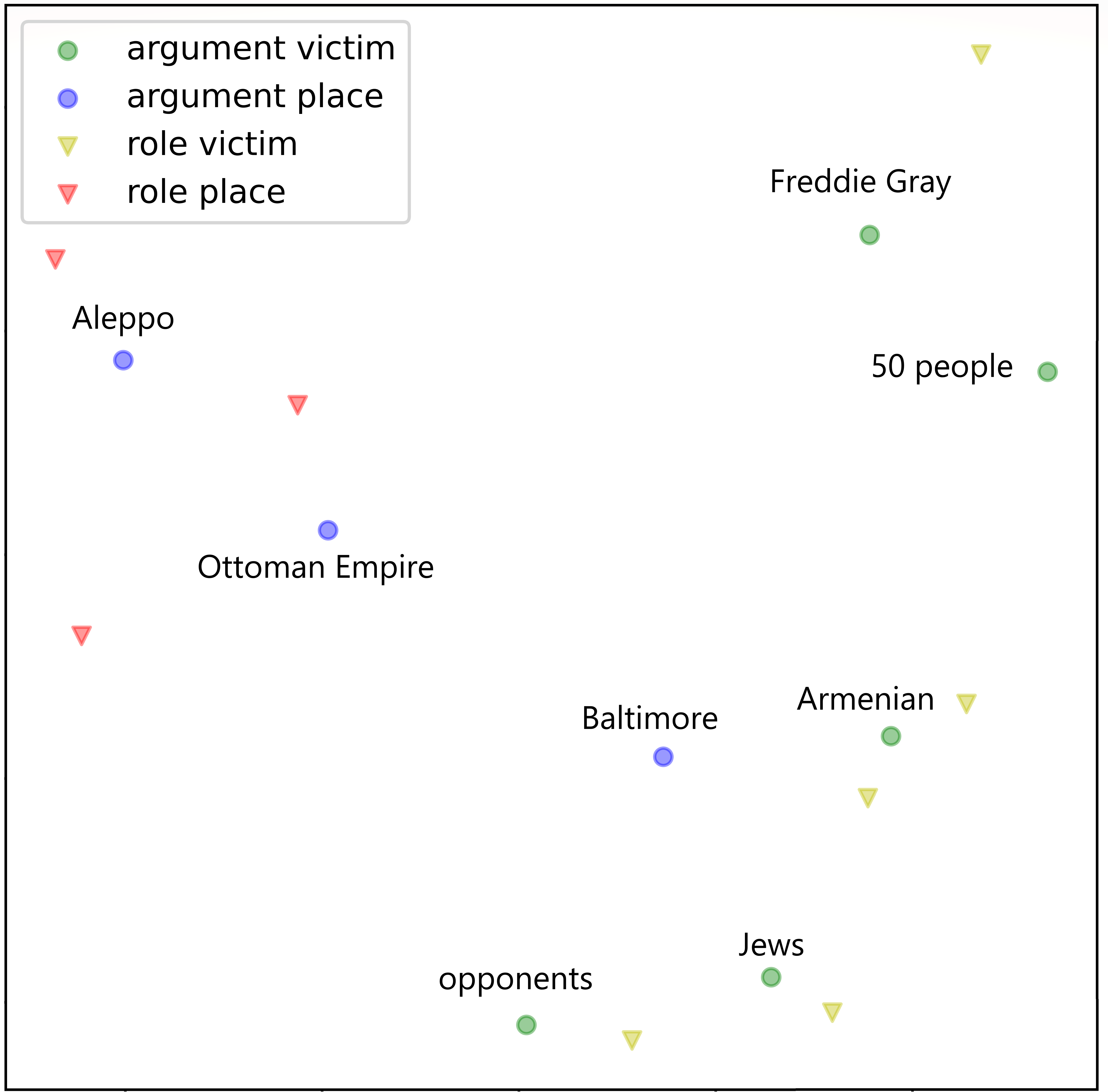}
    }
    \caption{A t-SNE visualization example from RAMS, where embeddings of arguments and roles are from 5 different documents. We use average pooling representations encoded by BERT for arguments in (a) and representations fused with latent role embeddings in (b). }
    \label{fig:fig4}
\end{figure}

Moreover, in order to verify the beneficial guidance of role representations,  we display the t-SNE~\cite{JMLR:v9:vandermaaten08a} visualization of arguments belonging to two different roles that co-occur in 5 different documents, along with corresponding latent role embeddings. As Figure~\ref{fig:c} shows, arguments belonging to the same role in different documents
are scattered over the whole embedding space due to their different target events and context. Notably, fused with latent role embeddings, in Figure~\ref{fig:d}, the representation of arguments belonging to \textit{victim} or \textit{place} is more adjacent, which illustrates our RLIG provides beneficial latent role information guidance. 



 \begin{table}[tbp]
 \setlength{\tabcolsep}{0.2mm}{
\begin{tabular}{llcccc}
\hline
\multirow{2}{*}{Method} &
  \multirow{2}{*}{Params} &
  \multicolumn{2}{c}{Dev} &
  \multicolumn{2}{c}{Test} \\ \cline{3-6} 
 &
   &
  \footnotesize{Span F1} &
  \footnotesize{Head F1} &
  \footnotesize{Span F1} &
  \footnotesize{Head F1} 
   \\ \hline
\footnotesize{${\text{SCPRG}}_{{\text{base}}}$}          & 119.37M            & 46.56      & 53.38      & 48.94       & 56.10                       \\
\quad \footnotesize{\textit{-STCP}} & 118.78M & 46.18      & 52.87      & 47.33       & 54.68                       \\
\quad  \footnotesize{\textit{-RLIG}}  & 118.78M & 45.27      & 52.36      & 47.91       & 54.52                       \\
\footnotesize{\textit{-STCP\&RLIG}}  & 118.19M & 45.07      & 51.59      & 45.76       & 53.16                       \\
\quad  \footnotesize{\textit{-ASE}} & 119.37M & 45.92      & 52.61      & 48.26       & 55.63                       \\ \hline
\footnotesize{${\text{SCPRG}}_{{\text{large}}}$}     & 372.90M               & 50.53      & 57.66      & 52.32       & 59.66                       \\
\quad  \footnotesize{\textit{-STCP}} & 371.85M & 49.94      & 56.55      & 50.89       & 57.57                       \\
\quad  \footnotesize{\textit{-RLIG}} & 371.68M & 49.96      & 57.32      & 51.28       & 58.46                       \\
\footnotesize{\textit{-STCP\&RLIG}}  & 370.63M & 48.33      & 54.04      & 47.52       & 55.61                       \\
\quad  \footnotesize{\textit{-ASE}}  & 372.90M & 49.80      & 56.31      & 51.73       & 58.48                       \\ \hline
\end{tabular}
}
\caption{Ablation Study on RAMS for ${\text{SCPRG}}$.}
\label{tab:table4}
\end{table}


\subsection{Analysis of Complexity and Compatibility}
SCPRG is a simple but effective framework for document-level EAE, where both STCP and RLIG introduce few parameters. Specifically, STCP leverages the well-learned attention heads from the pre-trained encoder and makes multiplication and normalization operation, which only introduces about 0.28\% new parameters as shown in Table~\ref{tab:table4}. Our RLIG only introduces about 0.3\% new parameters in the role embedding layer\footnote{ We add new special tokens for role types and therefore the RLIG module introduces more parameters in ${\text{SCPRG}}_{{\text{large}}}$.} and feature fusion layer. This makes the parameter quantity of our model approximate to the transformer-based encoder plus a MLP classifier.

Additionally, the two proposed techniques STCP and RLIG have good transportability, which can be easily applied to other event extraction models, leveraging the  attention heads of pre-trained transformer encoder such as BERT.

\section{Related Works}

\subsection{Sentence-level Event Extraction}
Previous approaches focus on extracting the event trigger and its arguments from a single sentence. \cite{chenetal2015event} firstly propose a neural pipeline model for event extraction and \cite{nguyenetal2016jointevent,nguyengrishman2015event,liuetal2017exploiting,zhou2020weighted} further extend the pipeline model to recurrent neural networks and convolutional neural networks. To model the dependency of words in a sentence,~\cite{liuetal2018jointly, yanetal2019event, fernandezastudilloetal2020transition} leverage dependency trees to model semantic and syntactic relations. \cite{waddenetal2019entity} enumerates
 all possible spans and construct span graphs with graph neural networks to propograte information. Some methods using transformer-based pre-trained model~\cite{waddenetal2019entity, wangetal2019adversarialtraining, tongetal2020improving, luetal2021text2event,liuetal2022dynamic} also achieve remarkable performance.

\subsection{Document-level Event Extraction}
In real-world scenarios,
a large number of event elements are expressed across sentences and therefore recent works begin to explore document-level event extraction (DEE).
DEE focuses on extracting event
arguments from an entire document and faces the challenge of the long distance dependency~\cite{wangetal2022query, xu-2022-xu}. 

For document-level EAE, the key step of DEE,
most of previous works mainly fall into three categories:
(1) tagging-based methods; (2) span-based methods; (3) generation-based methods. 
\cite{wangetal2021cleve, ducardie2020document} utilize the sequence labeling
model BiLSTM-CRF~\cite{zhangetal2015bidirectional} for DEE. \cite{zhengetal2019doc2edag} propose a transformer-based architecture and model DEE as a serial
prediction paradigm, where arguments are predicted in a predeﬁned role order. Base on their architecture,~\cite{xuetal2021document} construct a heterogeneous graph and a tracker module to capture the interdependency among events. However, tagging-based methods are inefficient due to the restriction to the extraction of individual
arguments, and the former extraction will not consider the latter extraction results. \cite{yangetal2021document} propose an encoder-decoder framework that extracts structured events in a parallel manner. Besides, \cite{renetal2022clio} integrate argument roles into document encoding to aware tokens of multiple role information for nested arguments problem. Other span based methods~\cite{ebner2020multi, zhang2020two} predict the argument roles for candidate text spans with a maximum length limitation. Moreover,~\cite{xu2022two} propose a two-stream encoder with AMR-guided graph to solve long-distance dependency problem.
On another aspect, ~\cite{lietal2021document} formulate the problem as conditional generation and~\cite{duetal2021grit} regards the problem as a sequence-to-sequence task. \cite{weietal2021trigger}  reformulate the task as reading a comprehension task. 


\section{Conclusion}
In this paper, we propose a novel SCPRG framework for document-level EAE that mainly consists of two compact, effective and transplantable  modules.
Specifically, our STCP adaptively aggregates the information of non-argument clue words and RLIG provides latent role 
information guidance containing semantic relevance among roles. Experimental results show that SCPRG outperforms
 existing state-of-the-art EAE models and further analyses demonstrate that our method is both
effective and explainable. For
future works, we hope to apply SCPRG to more information extraction tasks such
as relation extraction and multilingual extraction,
where contextual information plays a significant role.

\section{Limitations}
Although our experiments prove the superiority of our SCPRG model, it is only applicable to document-level EAE tasks with known event triggers because both STCP and RLIG calculate the attention product of the trigger and candidate spans. However, in real-life scenarios, event triggers are not always available. In view of this problem, we have a preliminary solution and plan to improve our model in the next work.  The core idea of our method is to select and integrate context and role information based on candidate arguments and target events. Based on this idea, we briefly provide two solutions for the above limitation. First, we can make the model predict the best candidate trigger words. Second, we can replace trigger words with special event tokens.  In the next work, we plan to extend our model to document-level EAE tasks without trigger words and evaluate it through extensive experiments.

\section*{Acknowledgements}
This work was supported by the National Science Foundation
of China under Grant 61976043, and in part by the Science and technology support program of Sichuan Province under Grant 2022YFG0313.

\bibliography{anthology,custom}
\bibliographystyle{acl_natbib}

\clearpage
\appendix

\begin{table}[]
 \centering
 \setlength{\tabcolsep}{0.7mm}{
\begin{tabular}{lcccc}
\hline
\multirow{2}{*}{\textbf{Method}} & \multicolumn{2}{c}{\textbf{Arg IF}} & \multicolumn{2}{c}{\textbf{Arg CF}} \\ \cline{2-5} 
                & \footnotesize{Head F1} & \footnotesize{Coref F1} & \footnotesize{Head F1} & \footnotesize{Coref F1} \\ \hline
${\text{SCPRG}}_{{\text{base}}}$      & 76.13   & 74.90   & 68.91   & 68.33   \\
\quad \textit{-STCP}           & 74.64   & 73.46   & 67.48   & 67.07  \\
\quad \textit{-RLIG}           & 75.59   & 73.83    & 68.26   & 67.49  \\
\quad \textit{-STCP\&RLIG}           & 73.92   & 73.22    & 66.90   & 65.98  \\
\quad \textit{-ASE} & 75.86        & 74.37        & 68.41        & 68.01       \\ \hline
${\text{SCPRG}}_{{\text{large}}}$      & 77.26   & 76.10   & 70.92   & 70.08   \\
\quad \textit{-STCP}           & 75.54   & 73.37    & 69.67   & 68.77   \\
\quad \textit{-RLIG}           & 76.30   & 74.08   & 69.87   & 68.96   \\
\quad \textit{-STCP\&RLIG}           & 75.45   & 73.55    & 68.63   & 67.30  \\
\quad \textit{-ASE} & 76.57        & 74.22        & 69.65        &  68.01       \\ \hline
\end{tabular}
}
\caption{Ablation Study on WikiEvent for ${\text{SCPRG}}$.}
\label{tab:table5}
\end{table}

\section{Ablation Study}
\label{sec:appendixB}
In the main body of the paper, we conduct ablation study on RAMS dataset for ${\text{SCPRG}}_{{\text{base}}}$ and ${\text{SCPRG}}_{{\text{large}}}$. In order to fully evaluate the effect of different components on our model, we also provide the results of the ablation study on WikiEvents for for ${\text{SCPRG}}_{{\text{base}}}$ and ${\text{SCPRG}}_{{\text{large}}}$. 

As shown in Table~\ref{tab:table5}, when we remove STCP module, both Head F1 and Coref F1 score of ${\text{SCPRG}}_{{\text{base}}}$ drop by \textbf{1.49/1.44} and
\textbf{1.43/1.26} on test set for argument identification task (Arg IF) and argument classification (Arg CF) task, which demonstrates that our STCP  captures the clue information of non-argument context that is significant for document-level EAE. 

Additionally, when removing RLIG module, the performance of ${\text{SCPRG}}_{{\text{large}}}$ drops sharply by \textbf{0.96/2.02} Head F1 and \textbf{1.05/1.12} Coref F1 on
Wikievent test set for both two tasks. 
Moreover, when we remove argument-impossible spans exclusion (ASE), both ${\text{SCPRG}}_{{\text{base}}}$ and ${\text{SCPRG}}_{{\text{large}}}$ have
a performance decay. These results indicate
that both STCP and ASE are beneficial. 
\begin{figure*} [htbp]
  \centering
  \includegraphics[width=1.0\linewidth]{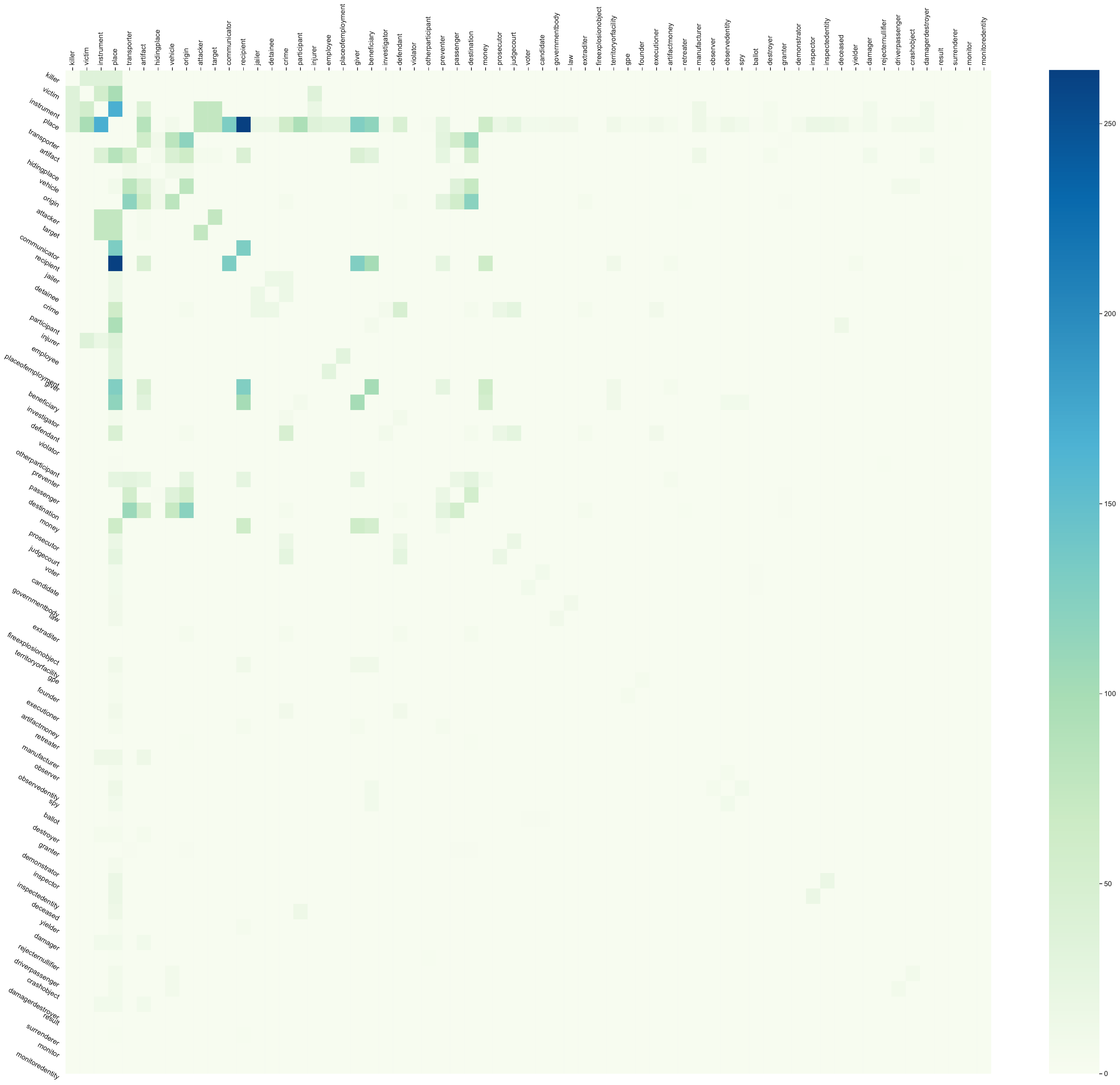} 
  \caption{Visualization of the co-occurrence frequency
between all roles in RAMS test set. we have reserved and set the co-occurrence number with itself to zero. } 
  \label{fig:fig7} 
\end{figure*} 

\section{Co-occurrence Frequency Matrix}
\label{sec:appendixC}
In this section, we show the complete co-occurrence frequency matrix 
which contains all roles in RAMS test set. We count the frequency of co-occurrence between every two roles and draw the heat map according to the frequency in Figure~\ref{fig:fig7}. 
It can be seen from the figure that the co-occurrence phenomenon exists between many roles, especially those occur in the same event, which indicates that there is semantic relevance among roles.

\end{document}